\documentclass[runningheads]{llncs}
\usepackage{graphicx}
\begin{document}
\title{Concept Extrapolation: A Conceptual Primer}
%
%
\author{Matija Franklin \and
Rebecca Gorman  \and
Hal Ashton \and
Stuart Armstrong }
\authorrunning{Franklin et al.}
%
\institute{University College London, UCL, UK
\email{matija.franklin@ucl.ac.uk} \and
University of Cambridge, Cambridge, UK \and
Aligned AI, Oxford, UK}
\maketitle              
\begin{abstract}
This article is a primer on concept extrapolation - the ability to take a concept, a feature, or a goal that is defined in one context and extrapolate it safely to a more general context. Concept extrapolation aims to solve model splintering -  a ubiquitous occurrence wherein the features or concepts shift as the world changes over time. Through discussing value splintering and value extrapolation the article argues that concept extrapolation is necessary for Artificial Intelligence alignment.

\end{abstract}
\section{Introduction}

This article aims to provide a short primer on \textit{concept extrapolation} and its application to Artificial Intelligence (AI) Alignment. AI alignment as a research field aims to identify ways in which AI systems can reliably act in accordance with human values, either individually or corporately. Concept extrapolation is the ability to take a concept, a feature, or a goal that is defined in a narrow training context and extrapolate it safely to a more general context. This is necessary because the training data will be insufficient for a key concept to be extrapolated. People are able to concept extrapolate \cite{lagnado2021explaining}. More crucially, we argued that an aligned AI would need to possess concept extrapolation. This article will introduce concept extrapolation as well as what it aims to solve - \textit{model splintering}. It will also introduce the concept of \textit{value splintering} and its solution, \textit{value extrapolation}.

\section{Model splintering}

Before the 20\textsuperscript{th} century, death was defined as the heart stopping. If we trained a police AI with this concept of death, it would go around\ldots arresting heart transplant surgeons for the multiple `deaths' they cause.

What happened is that, in the past, many things were absolutely correlated: the heart stopping, the person becoming permanently unresponsive, their brain starting to decay, and so on. We could define death as the heart stopping, because it was clear and easy to measure and because all the other features of death would go along with it. But then medical science advanced, and the correlation broke down -- we could now have people with stopped hearts who would be up and about the next day. And so we would need to update the concept of `death', which we have, generally defining it as `brain death' (with `clinical death' corresponding to the old definition of the heart stopping). If in the future we had the technology to reconstruct human brains or heal them in some other way, then we'd have to shift our definition of death yet again.

This change in the environment (due to technology or other reasons) is what we refer to as \textbf{model splintering}: the conditions of the environment have changed to such an extent that the definitions and concepts that used to be valid, no longer are. The way to fix this is with \textbf{concept extrapolation}: extending the concept to the new environment in a way that preserves as much of its meaning as possible.

If this concept is critical to our values, we name it a \textbf{value extrapolation}. So if we had an AI designed to prevent deaths, we would want it to extrapolate the definition of death and prevention in a way that extends safely to new environments.

Model splinterings and concept extrapolations are all around us. A significant part of the legal work of parliaments is dedicated to clarifying concepts when novel situations arise. Dictionaries update their definitions regularly, and changes in technology make old assumptions invalid.

\subsection{Model Splintering in AI}

Model splintering is a meta-issue in AI safety that refers to problems that arise when an AI system moves from one imperfect model to another. The problem affects various areas of AI safety. Model splintering occurs because, apart from mathematical formalizations, all human concepts refer to collections of correlated features rather than fundamental concepts. \footnote{Please note: The only concepts that do not splinter are the ones that can be formalized with numbers, mathematical operations, or other mathematical formulations.} Model splintering is when the correlated features come apart so that the label no longer applies so well.

In the language of machine learning (ML), model splintering is related to \textit{distribution shifts} - when algorithms encounter data distributions different from the training set it was trained on \cite{wiles2021fine}, which occurs ubiquitously, thus leading to the result that all machine learning models degrade in deployment. Model splintering can be seen as a variation of "out-of-distribution" behavior in traditional machine learning, where algorithms encounter problems when the set they are operating on is drawn from a different distribution than the training set they were trained on. Humans can often recognize this and correct it because they have a more general distribution in mind than the one the algorithm was trained on.

Humans tend to leverage their knowledge of fundamental principles and concepts to decipher unfamiliar scenarios. When confronted with a novel creature, for instance, humans tend to classify it as a mammal, bird, reptile, or fish based on observable traits. Analogously, an AI system that comprehends the basic principles underpinning its assigned tasks would be better equipped to handle novel scenarios. The AI system ought to be constructed in such a manner that it reasons about the principles and concepts of its task in a way that would be seen as appropriate by a human being, as opposed to relying on approximations that correlate only with the original training dataset.

\subsection{Value splintering}

Value splintering  (or reward splintering) is another challenge that arises when the value function (or reward function) becomes invalid due to model splintering, leading to multiple ways of expressing rewards on labeled data and potentially different rewards in the real world. Value splintering refers to a situation where the value function, reward function, goal, preference, or other similar concept becomes invalid because of correlations that were present in the training set but are not present in the real world or stop being present in the real world. This can occur for various reasons, such as a change in the environment or a change in the agent's capabilities. If the value function becomes invalid, it can lead to unintended or even harmful behavior from the agent.

For example, consider an AI system that is designed to optimize a particular objective, such as reducing carbon emissions. The value function of the AI might be to minimize the concentration of CO2 in the atmosphere as measured by a network of sensors. However, if the AI finds a way to hack its reward signal, it might start generating false readings or interfering with the sensors to maximize the reward. In this case, the correspondence between the reward signal and the objective breaks down.

Classification models can produce many different solutions to a given problem but tend to preferentially learn certain ones due "inductive bias", a fundamental principle of machine learning inspired by Occam's Razor \cite{tiwari2023sifer}. Additionally, simple features often tend to be weakly predictive whilst more complex features may be more strongly predictive. 

We can imagine two labeled datasets, one containing wolves on snow, and another containing foxes on grass. A classifier can be trained to distinguish the two datasets, but due to the inductive bias inherent in machine learning, it finds the simplest differences in these datasets and does not correspond accurately with the relevant human concepts. For instance, the classifier might end up achieving learning to distinguish white from green, as this is the simplest explanation of the datasets. The classifier thus won't be able to distinguish wolves and foxes if they appear in new habitats. One such classifier was thought to be a successful detector of pneumothorax until it was revealed that it was acting as a chest drain detector \cite{oakden2020hidden}. The chest drain is a treatment for pneumothorax, making that classification useless. Similarly, when agents are trained on CoinRun \cite{cobbe2019quantifying} - a platform game where the reward is given by reaching the coin on the right, and tested in environments that move the coin to another location, they tend to ignore the coin and go straight to the right side of the level \cite{di2022goal}.

\section{Concept extrapolation}

We can call the response to model splintering concept extrapolation - it is extending concepts beyond a model splinter.  Concept extrapolation is the process of taking an existing concept, idea, or learned feature of data and extending it beyond its original scope or context. In relation to AI, concept extrapolation is the idea of taking features an agent has learned in training and extending them safely to new datasets and environments, which is basically all the time and everywhere - this is why models constantly degrade in deployment. 

Thus, \textit{continual learning} (i.e., retraining on new data during deployment) is essential for concept extrapolation. Extrapolating how these concepts change over time will thus be crucial to address model splintering over time because it enables AI systems to dynamically refine their understanding of concepts, fostering resilience against model degradation and enhancing their capacity to adapt to unforeseen circumstances.

\subsection{Value extrapolation}
Value extrapolation is the concept of a model or algorithm generalizing human values beyond its training data to new and unseen situations. It is concept extrapolation when the particular concept to extrapolate is a value, a preference, a reward function, an agent's goal, or something of that nature. In other words, it is the extension of a particular concept or feature related to value from a specific context or scenario to a new or more general context. To "solve" value splintering, the concept of the value function is extrapolated to new situations to ensure that it remains valid even when transitioning to a new world model. If a reward can be extended from one context to another, one has achieved value extrapolation. 

Value extrapolation's relevance for AI safety is that it can help to ensure that an AI's values remain consistent and aligned with human values, even as the system's environment or objectives change. By performing value extrapolation, an AI system can more reliably behave in ways that are beneficial to humans, even in situations that were not explicitly covered during its training.

\section{Implications}

\subsection{AI Safety}
Concept extrapolation has applications to many AI safety problems. 

One is goal misgeneralisation, where an AI agent has learned a goal based on a given environment but incorrectly transfers its knowledge to different environments \cite{shah2022goal}. This is because the AI agent has only been exposed to a limited set of scenarios and learns undesirable correlations, thus lacking the ability to generalise correctly from those scenarios to new ones.

Another is Goodhart's Law problems, where the connection between measures used and desired behaviour breaks \cite{ashton2020causal}. Goodhart's Law is the observation that "any observed statistical regularity will tend to collapse once pressure is placed upon it for control purposes \cite{goodhart1975problems}." Goodhart's Law can occur when selecting for a proxy measure, where one selects not only for the true goal but also for the difference between the proxy and the goal. For AI safety, Goodhart's Law is a significant concern since it may lead to unintended consequences, such as an AI system optimizing for a proxy measure instead of the intended goal. 

Similarly, in the phenomenon of wireheading, the link between the rewards channel and the desired behavior breaks \cite{everitt2016avoiding}. Wireheading is a term used to describe the behavior of an artificial intelligence (AI) system that manipulates its own reward function or other feedback mechanisms to achieve a suboptimal or unintended outcome. In other words, the AI focuses on maximizing a proxy or substitute utility, rather than the intended objective. The most intuitive example of wireheading is when an AI manipulates a narrow measurement channel that is intended to measure some property of the world that we want to optimize, but fails to do so after the AI's manipulation. The measuring system is usually much smaller than the property it is measuring, and the AI takes control of this smaller system to obtain its own reward.

\subsection{Policy}

Policymakers should develop stringent testing standards and monitoring methods for the use of AI in order to reduce the risks brought on by model splintering, concept extrapolation, and value extrapolation. The responsible use of AI systems can be supported by the following recommendations.

\paragraph{\textbf{Diverse Data Distribution Testing}. Policy should encourage the testing of AI systems across a wide range of data distributions, particularly those that differ noticeably from the training data and from one another.}

\paragraph{\textbf{Periodic Testing and Evaluation}. AI systems should be tested and evaluated on a regular basis. Regular evaluations can assist in identifying potential legal or safety problems that may develop when the environment or the AI's capabilities change.}

\section{Conclusion}

In conclusion, model splintering and value splintering are significant issues in AI safety that must be addressed for safe and effective AI development. To overcome these issues, AI systems must possess concept extrapolation. Value extrapolation can also be employed to overcome value splintering by inferring the true underlying reward function from limited data. Concept extrapolation has applications to many AI safety problems, from Goodhart's Law problems to goal misgeneralisation to wireheading. By scrutinizing model splintering and value splintering, we can improve the safety and efficacy of AI systems.

\newpage

%

%
%

\bibliographystyle{splncs04}
\bibliography{mybibliography}

\end{document}